%% file: paper.tex
\documentclass[letterpaper]{article} 
\usepackage{aaai2026}
\usepackage{times}  
\usepackage{helvet}  
\usepackage{courier}  
\usepackage[hyphens]{url}  
\usepackage{graphicx} 
\usepackage{natbib}  
\usepackage{caption} 
\usepackage{algorithm}
\usepackage{algorithmic}
\usepackage{booktabs}

\usepackage{amsmath}
\usepackage{amsfonts}

\usepackage{newfloat}
\usepackage{listings}
\DeclareCaptionStyle{ruled}{labelfont=normalfont,labelsep=colon,strut=off} 
\floatstyle{ruled}
\newfloat{listing}{tb}{lst}{}
\floatname{listing}{Listing}
\nocopyright

\begin{document}

\title{ProbFM: Probabilistic Time Series Foundation Model with Uncertainty Decomposition}

\author{
    Arundeep Chinta\textsuperscript{\rm 1},
    Lucas Vinh Tran\textsuperscript{\rm 2},
    Jay Katukuri\textsuperscript{\rm 1}
}
\affiliations{
    \textsuperscript{\rm 1}JPMorganChase, Palo Alto, USA\\
    \textsuperscript{\rm 2}JPMorganChase, London, UK\\
    arundeep.chinta@jpmchase.com, lucas.vinhtran@chase.com, jay.katukuri@chase.com
}

\maketitle

\begin{abstract}
\input{abstract}

\end{abstract}

\section{Introduction}\label{sec:introduction}
\input{0.introduction}

\section{Methodology: ProbFM Architecture and Training}\label{sec:methodologyProbFM}
\input{1.methodology}

\section{Experimental Setup}\label{sec:experiments}
\input{2.experiments}

\section{Results and Analysis}\label{sec:results}
\input{3.results}

\section{Conclusion}\label{sec:conclusion}
\input{6.conclusion}

\clearpage
\section{Disclaimer}\label{sec:disclaimer}
\input{7.disclaimer}

\bibliography{aaai2026}

\appendix
\input{8.appendix}

\end{document}

%% file: abstract.tex
Time Series Foundation Models (TSFMs) have emerged as a promising approach for zero-shot financial forecasting, demonstrating strong transferability and data efficiency gains. However, their adoption in financial applications is hindered by fundamental limitations in uncertainty quantification: current approaches either rely on restrictive distributional assumptions, conflate different sources of uncertainty, or lack principled calibration mechanisms. While recent TSFMs employ sophisticated techniques such as mixture models, Student's t-distributions, or conformal prediction, they fail to address the core challenge of providing theoretically-grounded uncertainty decomposition. 

For the very first time, we present a novel transformer-based probabilistic framework, \textbf{ProbFM} (\textit{probabilistic foundation model}), that leverages Deep Evidential Regression (DER) to provide principled uncertainty quantification with explicit epistemic-aleatoric decomposition. Unlike existing approaches that pre-specify distributional forms or require sampling-based inference, ProbFM learns optimal uncertainty representations through higher-order evidence learning while maintaining single-pass computational efficiency. 

To rigorously evaluate the core DER uncertainty quantification approach independent of architectural complexity, we conduct an extensive controlled comparison study using a consistent LSTM architecture across five probabilistic methods: DER, Gaussian NLL, Student's-t NLL, Quantile Loss, 
and Conformal Prediction. This design isolates the contribution of different uncertainty quantification strategies while maintaining architectural consistency, providing clean evidence for DER's effectiveness before integration into more complex architectures.

Evaluation on cryptocurrency return forecasting demonstrates that DER maintains competitive forecasting accuracy while providing explicit epistemic-aleatoric uncertainty decomposition. We demonstrate practical value through uncertainty-aware trading strategies, showing how epistemic-aleatoric decomposition enables effective risk management by filtering high-uncertainty predictions.

This work establishes both an extensible framework for principled uncertainty quantification in foundation models and empirical evidence for DER's effectiveness in financial applications.

%% file: 0.introduction.tex
The emergence of Time Series Foundation Models (TSFMs) has advanced forecasting capabilities across domains, but a critical gap remains in uncertainty quantification for high-stakes applications. Current TSFMs—despite employing sophisticated techniques—fail to provide the principled uncertainty decomposition and calibration guarantees required for effective decision-making in quantitative finance. Current state-of-the-art TSFMs employ various sophisticated techniques. For example, Toto \citep{cohen2024toto} uses Student-T Mixture Models for observability data, MOIRAI \citep{woo2024moirai} implements four-component flexible mixtures for universal forecasting, Lag-Llama \citep{rasul2023lagllama} applies Student's $t$-distributions with foundation model training, and TimeGPT \citep{garza2023timegpt} leverages conformal prediction for coverage guarantees. While these approaches achieve strong empirical performance, they share fundamental limitations that constrain their applicability in scenarios where understanding the source and reliability of uncertainty is crucial, especially in quantitative financial applications.

The core problem lies in the conflation of different uncertainty sources and the lack of principled frameworks for learning optimal uncertainty representations. Mixture model approaches, while empirically successful, pre-specify distributional components and cannot distinguish between epistemic uncertainty (reducible through more data) and aleatoric uncertainty (irreducible data noise). Parametric approaches like Student's $t$-distributions impose strong distributional assumptions that may not hold across diverse time series characteristics. Conformal prediction provides coverage guarantees but operates post-hoc without integration into the learning process, limiting its ability to provide nuanced uncertainty information for decision-making. Moreover, architectural differences across foundation models make it difficult to isolate which improvements stem from uncertainty quantification strategies versus architectural choices. These limitations become particularly critical in financial applications where position sizing must account for different uncertainty sources, and risk management depends on understanding whether prediction uncertainty reflects insufficient data or inherent market randomness.

We address these limitations through a two-part approach that combines architectural innovation with rigorous empirical validation. We introduce ProbFM (\textit{probabilistic foundation model}), a transformer-based probabilistic framework that integrates Deep Evidential Regression (DER) \citep{sensoy2018evidential, amini2020deep} with Normal-Inverse-Gamma (NIG) priors. This marks the 
\textit{first} application of evidential regression to time series foundation model architectures. Unlike existing probabilistic time series models that either fix distributional families a priori or require computationally expensive sampling, ProbFM learns to estimate uncertainty over distribution parameters themselves, enabling principled epistemic-aleatoric decomposition in a single forward pass.

However, the proliferation of different architectures in recent foundation models—from MOIRAI's mixture-based universal forecasting to Lag-Llama's Student-t parameterization—makes it challenging to assess whether performance gains stem from superior uncertainty quantification or simply from better architectures. To isolate the contribution of DER independent of architectural complexity, we conduct a controlled comparison using a consistent LSTM backbone across five probabilistic approaches. This methodological choice deliberately prioritizes internal validity: by holding architecture constant while varying only the uncertainty quantification mechanism, we obtain clean evidence for DER's effectiveness that generalizes beyond any specific model design.

Our evaluation demonstrates both theoretical promise and practical utility. On cryptocurrency return forecasting, DER-based models achieve competitive point prediction accuracy while providing explicit epistemic-aleatoric uncertainty decomposition. More importantly, we validate the decision-making value of this uncertainty decomposition through two financial applications. First, uncertainty-aware trade filtering—which selectively executes trades based on epistemic and aleatoric uncertainty thresholds—improves risk-adjusted returns by avoiding positions during high-uncertainty regimes. Second, Portfolio optimization using uncertainty-based position sizing demonstrates practical benefits, outperforming equal-weighted baselines by dynamically adjusting capital allocation based on prediction confidence.

All in all, our key contributions are:
\begin{itemize}
    \item \textbf{First application of DER to TSFMs}: To the best of our knowledge, we present the first application of DER with NIG priors to TSFMs, adapting the deep evidential regression approach of \citet{amini2020deep} to handle sequential time series data with explicit epistemic-aleatoric uncertainty decomposition through principled Bayesian inference.
    
    \item \textbf{Integrated coverage optimization}: We integrate a coverage loss mechanism to ensure prediction interval reliability, complementing the standard DER training objective with direct coverage probability optimization without requiring post-hoc calibration or arbitrary binning strategies.

    \item \textbf{Evidence annealing schedule}: We propose an evidence annealing schedule that directly controls evidence accumulation during training, building on the regularization annealing approach of \citet{sensoy2018evidential} to prevent overconfidence in early training phases.
    
    \item \textbf{Controlled empirical evaluation}: We conduct rigorous comparative evaluation using established uncertainty quantification metrics on a consistent architecture, systematically isolating the impact of uncertainty quantification approaches independent of architectural differences.
    
    \item\textbf{Financial decision-making validation}: We demonstrate practical utility through uncertainty-aware trading, showing how epistemic-aleatoric decomposition provides actionable
insights
\end{itemize}

%% file: 1.methodology.tex
Figure \ref{fig:probfm_architecture} in the Appendix~\ref{appendix:pipeline} presents the complete methodology pipeline and details the architectural flow diagram of our proposed ProbFM. Specifically, ProbFM provides a straightforward application of Deep Evidential Regression (DER) to single-variate single-step time series forecasting with six key components: (1) input processing with adaptive patching and positional encoding, (2) standard transformer architecture for representation learning, (3) DER head for principled uncertainty estimation and epistemic-aleatoric decomposition, (4) combined loss functions including evidential and coverage losses, (5) single-stage training with AdamW optimization \citep{loshchilov2017adamw} and evidence annealing for simplicity and efficiency, and (6) single-pass inference providing point predictions, uncertainty decomposition, and confidence intervals.

In this section, we begin by formally defining the problem and establishing the theoretical underpinnings of our method. Subsequently, we introduce the design of the ProbFM architecture, which firstly encompasses input encoding and patch-based processing, a transformer backbone, and a Deep Evidential Regression head. Furthermore, we propose an improved loss function that incorporates coverage mechanisms. The section also covers the training procedure and optimization strategies, followed by the inference process and uncertainty quantification, which completes the diagram flow as illustrated in Figure \ref{fig:probfm_architecture}.

\subsection{Problem Formulation and Theoretical Foundation}\label{sec:prob_formulation}

Consider a univariate time series dataset $\mathcal{D} = \{(x_i, y_i)\}_{i=1}^N$ where $x_i \in \mathbb{R}^T$ represents the input context window of length $T$ and $y_i \in \mathbb{R}$ represents the single-step target forecast. Traditional probabilistic approaches learn a mapping $f_\theta: x \rightarrow p(y|x)$ by parameterizing a specific distribution family. In contrast, Deep Evidential Regression (DER) \citep{sensoy2018evidential, amini2020deep} learns a mapping $f_\theta: x \rightarrow p(\phi|x)$ where $\phi$ represents the parameters of the predictive distribution, enabling uncertainty about the distribution parameters themselves.

For single-step univariate time series forecasting, we employ a Normal-Inverse-Gamma (NIG) prior \citep{gelman2013bayesian} to model uncertainty over Gaussian distribution parameters:
\begin{equation}
p(\mu, \sigma^2) = \text{NIG}(\mu, \lambda, \alpha, \beta)
\end{equation}
in which the NIG distribution is factorized as:
\begin{align}
p(\mu|\sigma^2) &= \mathcal{N}(\mu, \sigma^2/\lambda) \\
p(\sigma^2) &= \text{Inverse-Gamma}(\alpha, \beta)
\end{align}
where $\mu, \lambda, \alpha, \beta$, ... represent distribution parameters.

This formulation enables explicit decomposition of total uncertainty into epistemic and aleatoric components:
\begin{align}
\mathbb{E}[y|x] &= \mu \\
\text{Aleatoric Uncertainty} &= \mathbb{E}[\sigma^2|x] = \frac{\beta}{\alpha - 1} \\
\text{Epistemic Uncertainty} &= \text{Var}[\mu|x] = \frac{\beta}{(\alpha - 1)\lambda}
\end{align}

Notably, our framework can be readily generalized to multi-horizon and multivariate forecasting by employing a Normal-Inverse-Wishart (NIW) prior \citep{gelman2013bayesian} instead of NIG, where the mean becomes a vector $\mu \in \mathbb{R}^H$ for H-step ahead forecasting, and the covariance becomes a matrix $\Sigma \in \mathbb{R}^{H \times H}$ to capture dependencies across time steps and variables. In this paper, we decide to exploit NIG because establishing the effectiveness of DER for the fundamental single-variate, single-step case provides a solid foundation before extending to more complex multi-horizon, multivariate scenarios in future work.

\subsection{ProbFM Architecture Design}

\subsubsection{Input Encoding and Patch-Based Processing}

ProbFM employs a sophisticated input processing pipeline designed for efficient handling of univariate time series across different frequencies and scales. The architecture begins with patch-based processing following PatchTST \citep{nie2022patchtst}, enhanced with an adaptive multi-patch size approach inspired by MOIRAI \citep{woo2024moirai} that adjusts patch size based on input characteristics; followed by Patch Embedding with Positional Encoding.

\textbf{Adaptive Patching Strategy:}
The patch size selection follows the frequency-based mapping proposed by \citet{woo2024moirai}:
\begin{equation}
\small
\text{patch\_size} = \begin{cases}
8 & \text{if frequency} \in \{\text{yearly, quarterly}\} \\
8, 16, 32 & \text{if frequency} \in \{\text{monthly}\} \\
16, 32 & \text{if frequency} \in \{\text{weekly, daily}\} \\
32, 64 & \text{if frequency} \in \{\text{hourly}\} \\
32, 64, 128 & \text{if frequency} \in \{\text{minute-level}\} \\
64, 128 & \text{if frequency} \in \{\text{second-level}\}
\end{cases}
\end{equation}

This multi-patch size framework enables efficient handling of time series spanning diverse frequencies within a unified architecture, automatically adapting the temporal granularity based on input characteristics.

\textbf{Patch Embedding with Positional Encoding (PE):}
\begin{align}
X_{\text{patches}} &= \text{Patching}(X_{\text{raw}}, \text{patch\_size}) \\
X_{\text{encoded}} &= \text{LinearProjection}(X_{\text{patches}}) + \text{PositionalEncoding}
\end{align}

Moreover, our positional encoding also incorporates temporal information following \citet{vaswani2017attention}:
\begin{equation}
\text{PE}_{(t,d)} = \begin{cases}
\sin(t/10000^{d/d_{model}}) & \text{if } d \text{ is even} \\
\cos(t/10000^{d/d_{model}}) & \text{if } d \text{ is odd}
\end{cases}
\end{equation}
where $t$ is the position in the sequence, $d$ is the dimension index, and $d_{\text{model}}$ is the total embedding dimension.

\subsubsection{Transformer Architecture}

After obtaining patch embeddings with Positional Encoding, ProbFM then employs a standard transformer architecture \citep{vaswani2017attention} for learning representations from time series data:

\textbf{Multi-Head Attention:}
\begin{equation}
\text{Attention}(Q, K, V) = \text{softmax}\left(\frac{QK^T}{\sqrt{d_k}}\right)V
\end{equation}

\textbf{Feed-Forward Network:}
\begin{equation}
\text{FFN}(x) = \text{SwiGLU}(W_1 x + b_1)
\end{equation}
where SwiGLU is the gated linear unit activation function introduced by \citet{shazeer2020glu}.

\subsubsection{Deep Evidential Regression Head}\label{sec:der_head}

Lastly, we implement the DER head, in which it maps transformer outputs directly to NIG distribution parameters following the parameter projection approach of \citet{amini2020deep}:

\textbf{NIG Parameter Projection:}
\begin{align}
\small
\mu &= W_\mu h + b_\mu \\
\lambda &= \text{Softplus}(W_\lambda h + b_\lambda) + \epsilon \\
\alpha &= \text{Softplus}(W_\alpha h + b_\alpha) + 1 + \epsilon \\
\beta &= \text{Softplus}(W_\beta h + b_\beta) + \epsilon
\end{align}

These constraints are essential to ensure the mathematical validity of the Normal-Inverse-Gamma distribution. The NIG distribution requires specific parameter domains: $\mu \in \mathbb{R}$, $\lambda > 0$, $\alpha > 1$, and $\beta > 0$. The Softplus activation function ensures non-negative outputs, while the additive constants guarantee strict positivity where required. Specifically, $\mu$ requires no constraint as it represents the mean parameter, $\lambda$ (precision parameter) must be positive for the normal distribution, $\alpha$ must exceed 1 for the inverse-gamma distribution to be well-defined, and $\beta$ (scale parameter) must be positive. Without these constraints, the neural network could output invalid parameter values leading to undefined probability distributions and numerical instabilities during training and inference.

To the best of our knowledge, this work represents the first application of Deep Evidential 
Regression with Normal-Inverse-Gamma priors to time series foundation model 
architectures. While DER has been successfully applied to 
regression tasks \citep{amini2020deep}, prior work has not explored its integration with TSFMs
for sequential time series forecasting. This 
integration addresses three fundamental limitations in existing probabilistic 
time series foundation models:

(1) \textbf{Distributional Flexibility Without Pre-specification}: Unlike parametric 
approaches (e.g., Student's t-distribution in Lag-Llama \citep{rasul2023lagllama}) that fix the 
distributional family a priori, or mixture models (e.g., MOIRAI \citep{woo2024moirai}) that 
pre-specify mixture components, DER learns a distribution over distribution 
parameters through higher-order evidence learning. While the NIG form itself 
is specified, the model learns optimal uncertainty representations rather than 
point estimates of fixed distribution parameters.

(2) \textbf{Explicit Uncertainty Decomposition}: Unlike existing TSFMs that conflate 
different sources of uncertainty or provide only total predictive variance, DER 
enables principled decomposition into epistemic uncertainty (reducible through 
more data) and aleatoric uncertainty (irreducible data noise). This decomposition 
is derived directly from the NIG posterior without requiring post-hoc analysis 
or additional calibration procedures.

(3) \textbf{Single-Pass Computational Efficiency}: Unlike mixture model approaches 
that require sampling from multiple components \citep{woo2024moirai} or conformal prediction 
methods \citep{garza2023timegpt} that require calibration sets, DER provides complete uncertainty 
quantification—including point predictions, uncertainty decomposition, and 
confidence intervals—in a single forward pass through the network.

These capabilities directly enable the uncertainty-aware decision-making 
applications demonstrated in our financial experiments in Section \ref{sec:results}.

\subsection{Enhanced Loss Function via Coverage Integration}

\subsubsection{Evidential Loss Function.}
We propose the core evidential loss following Amini et al.~\cite{amini2020deep} which encourages the model to place evidence proportional to prediction accuracy:
\begin{equation}
\mathcal{L}_{\text{EDL}} = \mathcal{L}_{\text{NLL}} + \lambda_{\text{evd}} \mathcal{L}_{\text{reg}}
\end{equation}

This can be further expanded as:
\begin{equation}
\small
\begin{aligned}
\mathcal{L}_{\text{NLL}} = &\frac{1}{2}\log\left(\frac{\pi}{\lambda}\right) - \alpha \log(\Omega) \\
&+ \left(\alpha + \frac{1}{2}\right) \log\left(\frac{(y-\mu)^2}{\lambda} + \Omega\right) \\
&+ \log\left(\frac{\Gamma(\alpha)}{\Gamma(\alpha + \frac{1}{2})}\right)
\end{aligned}
\end{equation}
where $\Omega = 2\beta(1+\lambda)$ and $D_{\text{post}}$ represents the posterior NIG distribution after incorporating evidence from the neural network.

Our Regularization Term is defined as:
\begin{equation}
\mathcal{L}_{\text{reg}} = |\hat{y} - y| \cdot (\alpha + \nu - 2)
\end{equation}
where this regularization penalizes high evidence (low uncertainty) when predictions are inaccurate, encouraging proper uncertainty calibration.

\subsubsection{Coverage Loss Component.}\label{sec:coverage_loss}
To ensure that prediction intervals contain the true values at specified confidence levels, we employ a Prediction Interval Coverage Loss:
\begin{equation}
\mathcal{L}_{\text{coverage}} = |\text{PICP}_{\text{target}} - \text{PICP}_{\text{actual}}|
\end{equation}
where $\text{PICP}_{\text{target}}$ is the desired coverage probability (e.g., 0.95 for 95\% confidence intervals) and $\text{PICP}_{\text{actual}}$ is the actual coverage probability computed as:
\begin{equation}
\text{PICP}_{\text{actual}} = \frac{1}{N}\sum_{i=1}^{N} \mathbb{I}(y_i \in [\text{CI}_{\text{lower},i}, \text{CI}_{\text{upper},i}])
\end{equation}
where $\mathbb{I}(\cdot)$ is the indicator function that equals 1 when the true value $y_i$ falls within the predicted confidence interval $[\text{CI}_{\text{lower},i}, \text{CI}_{\text{upper},i}]$.

\noindent\textbf{Integrated Coverage Optimization:} While coverage-based loss functions exist for prediction interval estimation,
to the best of our knowledge, this work represents the first integration of coverage loss with Deep Evidential Regression for time series forecasting (Equation~(\ref{eq:combined_loss})). Unlike calibration approaches such as Expected Calibration Error (ECE)~\cite{naeini2015obtaining} that require arbitrary choices of binning strategies or temperature scaling methods~\cite{guo2017calibration} that introduce additional tunable parameters, our coverage loss directly optimizes the target metric (prediction interval coverage probability) without additional hyperparameters beyond the desired confidence level.

\subsubsection{Combined Training Objective.}
The final training objective combines the evidential loss with the coverage loss as: 
\begin{equation}
\label{eq:combined_loss}
\mathcal{L}_{\text{ProbFM}} = \mathcal{L}_{\text{EDL}} + \lambda_{\text{coverage}} \cdot \mathcal{L}_{\text{coverage}} + \lambda_{\text{wd}} ||\theta||^2_2
\end{equation}
where $\lambda_{\text{coverage}}$ is a fixed hyperparameter that controls the importance of coverage accuracy relative to the evidential loss, and $\lambda_{\text{wd}}$ is the $L_2$ regularization coefficient that controls weight decay strength to prevent overfitting. The combined objective thus balances forecasting accuracy, uncertainty calibration, and model generalization. These hyperparameters can be tuned via cross-validation for specific applications.

\subsection{Training Procedure and Optimization}

\subsubsection{Single-Stage Training Protocol}

ProbFM employs a straightforward single-stage training approach:

\textbf{Training Procedure:}
\begin{itemize}
\item Joint optimization of both forecasting accuracy and coverage loss from the start
\item Learning rate: standard cosine annealing with warmup
\item Fixed $\lambda_{\text{coverage}}$ weight throughout training
\end{itemize}

\subsubsection{Optimization Strategy} 

We employ AdamW with gradient clipping and learning rate scheduling with cosine annealing for our optimization strategy.

\textbf{AdamW with Gradient Clipping} \citep{loshchilov2017adamw}:
\begin{equation}
\theta_{t+1} = \theta_t - \eta \cdot \text{clip}(\nabla_\theta \mathcal{L}, \text{max\_norm})
\end{equation}
where gradient clipping with max\_norm $\in [1.0, 5.0]$ prevents gradient explosion, which is particularly important in evidential regression training due to the sensitivity of NIG parameter updates.

\textbf{Learning Rate Scheduling with cosine annealing} \citep{loshchilov2016sgdr}:
\begin{equation}
\eta_t = \eta_{\text{base}} \cdot \cos\left(\frac{\pi t}{T_{\text{total}}}\right) \cdot \text{warmup}(t)
\end{equation}
where warmup linearly increases the learning rate during an initial warmup period (typically 5-15\% of total training steps) to stabilize early-stage evidence accumulation. Moreover, building on the annealing approaches in EDL training \citep{sensoy2018evidential}, we further propose an evidence annealing schedule to prevent overconfidence during DER training:

\textbf{Evidence Annealing:}
\begin{equation}
\text{evidence\_scale}(t) = \min(1.0, t/T_{\text{anneal}})
\end{equation}

While Sensoy et al. \citep{sensoy2018evidential} employ KL regularization annealing that gradually increases the weight of the ``I do not know'' penalty term, our evidence annealing directly scales the evidence contribution during optimization. This annealing factor starts at 0 and linearly increases to 1.0 over $T_{\text{anneal}}$ training steps (typically 10-20\% of total training), preventing the model from placing excessive confidence in early predictions when the learned representations are still unstable. Unlike regularization weight annealing, this approach directly controls the evidence accumulation process, helping prevent evidence collapse—a common issue in EDL where the model generates overly confident predictions before meaningful patterns are learned.

\subsection{Inference and Uncertainty Quantification}

At inference time, ProbFM performs a single forward pass through the network to obtain the four NIG distribution parameters ($\mu, \lambda, \alpha, \beta$) from the DER head (Section \ref{sec:der_head}). These parameters fully characterize the posterior predictive distribution, from which we derive all uncertainty quantities and confidence intervals without additional sampling or calibration.

\subsubsection{Single-Pass Uncertainty Estimation}

ProbFM provides complete uncertainty quantification in a single forward pass based on explicit decomposition as described in Section \ref{sec:der_head}.
\textbf{Predictive Mean:}
\begin{equation}
\hat{y} = \mu
\end{equation}

\textbf{Total Predictive Variance:}
\begin{equation}
\text{Var}[y|x] = \frac{\beta}{\alpha-1} \left(1 + \frac{1}{\lambda}\right)
\end{equation}

\textbf{Epistemic Uncertainty (Model Uncertainty):}
\begin{equation}
\mathbb{U}_{\text{epistemic}} = \frac{\beta}{(\alpha-1)\lambda}
\end{equation}

\textbf{Aleatoric Uncertainty (Data Noise):}
\begin{equation}
\mathbb{U}_{\text{aleatoric}} = \frac{\beta}{\alpha-1}
\end{equation}

\subsubsection{Confidence Interval Construction.}

For prediction intervals at confidence level $1-\alpha$, ProbFM uses the 
posterior predictive distribution following the Deep Evidential Regression 
framework \cite{amini2020deep, sensoy2018evidential}. Under the NIG prior, 
the predictive distribution is a Student-t, yielding confidence intervals as:

\begin{equation}
\text{CI}_{1-\alpha} = \hat{y} \pm t_{2\alpha,\alpha/2} \sqrt{\frac{\beta(\lambda+1)}{\alpha\lambda}}
\end{equation}

where $t_{2\alpha,\alpha/2}$ is the critical value from the Student-t distribution 
with $2\alpha$ degrees of freedom.

%% file: 2.experiments.tex
\subsection{Dataset}

We conduct experiments on a Cryptocurrency Dataset consisting of the 11 most liquid cryptocurrency assets downloaded from Stooq.\footnote{\url{https://stooq.com}}
The dataset spans from January 1, 2020 to October 3, 2025, though not all assets have complete historical data extending back to the start date due to varying launch dates and market availability.

Our \textbf{Cryptocurrency Dataset (Daily Returns)}:
\begin{itemize}
\item \textbf{Assets}: 11 most liquid cryptocurrencies including ADA, BNB, BTC, DASH, DOGE, ETH, LTC, SOL, USDC, USDT, and XRP
\item \textbf{Time Period}: January 1, 2020 to October 3, 2025 (subject to asset availability)
\item \textbf{Construction}: 1-day log returns calculated as $r_t = \log(P_t/P_{t-1})$ where $P_t$ represents the closing price at time $t$
\end{itemize}

\begin{figure}[htbp]
    \centering
    \includegraphics[width=\columnwidth]{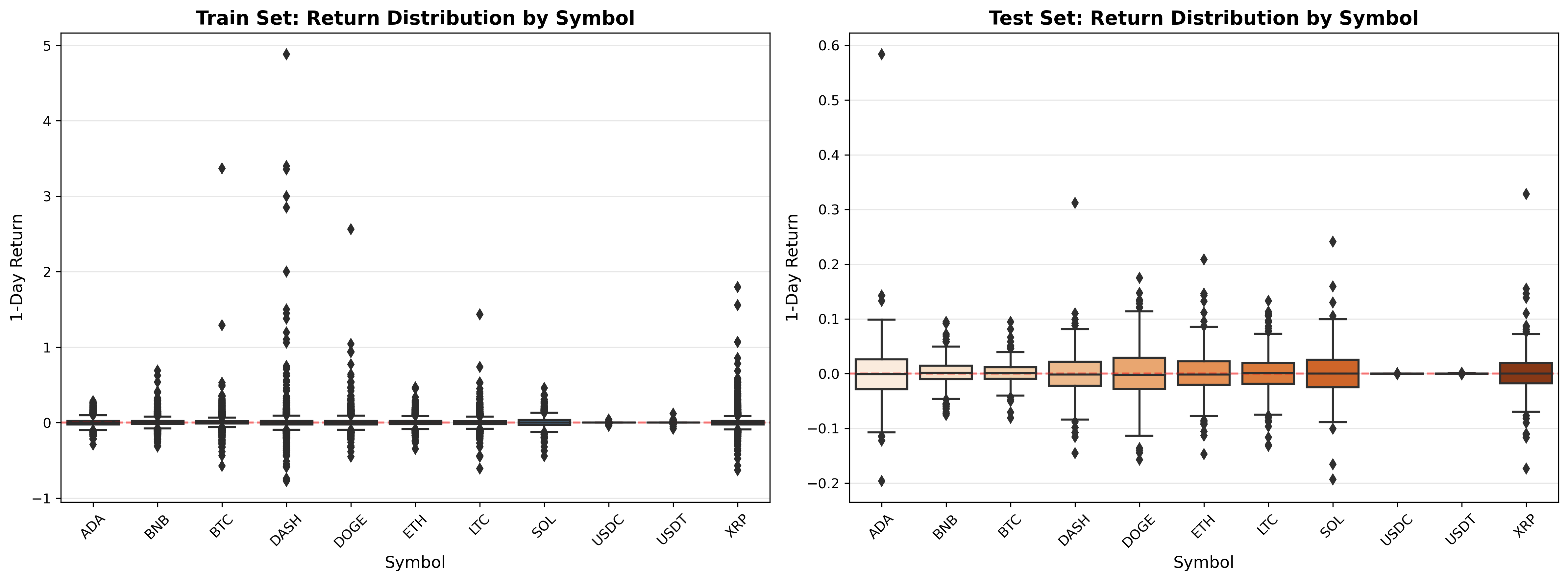}
    \caption{1-day log return distribution across assets in the cryptocurrency dataset. Left panel shows the training set distribution, right panel shows the test set distribution 
    .}
    \label{fig:return_distribution_boxplot}
    \vspace{-3ex}
\end{figure}

\subsection{Controlled Empirical Evaluation}

We compare the evidential regression approach proposed in ProbFM against other probabilistic forecasting methods. To ensure fair comparison, all methods use the same base model architecture: a 1-layer LSTM with 32 hidden dimensions. This eliminates the need for extensive hyperparameter tuning across different architectures and isolates the impact of different loss functions and uncertainty quantification approaches. Table \ref{tab:loss_comparison} summarizes the different loss functions and their probabilistic modeling capabilities evaluated in our study.

\begin{table}[h]
\caption{Comparison of Loss Functions and Probabilistic Modeling Approaches}
\vspace{-2ex}
\label{tab:loss_comparison}
\begin{center}
\resizebox{\linewidth}{!}{%
\begin{tabular}{lccc}
\toprule
\textbf{Loss Function} & \textbf{Probabilistic} & \multicolumn{2}{c}{\textbf{Representative Models}} \\
\cmidrule(lr){3-4}
 & & \textbf{Deep Learning} & \textbf{Foundation Models} \\
\midrule
MSE & No & Standard LSTM/GRU & - \\
Huber & No & Robust Regression Models & - \\
Gaussian NLL & Yes & DeepAR, LSTM-NLL & - \\
Student-t NLL & Yes & Student-t RNN & Lag-Llama \\
Quantile Loss & Yes & TFT, DeepQuantile & - \\
Mixture of Distributions & Yes & MDN-RNN, MixtureDensityLSTM & MOIRAI, Toto\\
Conformal Prediction & Yes & CP-LSTM & TimeGPT \\
\textbf{Evidential Regression} & Yes & Evidential LSTM/GRU & \textbf{ProbFM (Ours)} \\
\bottomrule
\end{tabular}}
\end{center}
\end{table}

The reason we choose this setup is to conduct a controlled evaluation 
that isolates the contribution of uncertainty quantification strategies 
independent of architectural differences. Recent foundation models employ 
diverse architectures,
making it difficult to disentangle whether performance gains 
stem from superior uncertainty quantification or architectural choices. By 
holding the architecture constant while varying only the probabilistic modeling 
approach, any observed differences in uncertainty calibration or predictive 
performance can be directly attributed to the loss function and uncertainty 
estimation mechanism rather than confounding architectural factors.

All methods are trained using the same optimization procedure, learning rate schedule, and training epochs. The only differences are the loss function and the output head that produces uncertainty estimates.

\subsection{Evaluation Metrics}

Our evaluation framework encompasses three complementary categories of metrics to provide a comprehensive assessment of model performance: accuracy metrics, probabilistic metrics, and trading-specific metrics as described in Appendix.

\subsubsection{Accuracy Metrics}

We evaluate point prediction accuracy using standard regression metrics

\subsubsection{Probabilistic Metrics}

To assess the quality of uncertainty estimates, we employ several probabilistic evaluation metrics 

\subsubsection{Trading Metrics}

We evaluate practical trading performance using risk-adjusted return metrics 

\subsection{Experimental Protocol}

\textbf{Data Splitting:} Please refer to Appendix 

\noindent\textbf{Preprocessing Pipeline:} Please refer to Appendix.
After model inference, predictions and uncertainty estimates are denormalized to recover values in the original return scale using the per-symbol statistics from training data.

\noindent\textbf{Sequence Generation:} For each symbol, we create sequences of length 50 (lookback window) from the normalized return data. The target for each sequence is the 1-day return at the end of the sequence, scaled to percentage (multiplied by 100). Sequences are generated separately per symbol and then combined while maintaining symbol identity for proper threshold application and evaluation.

\noindent\textbf{Model Architecture:} All methods use a consistent architecture for fair comparison: a 1-layer LSTM with 32 hidden dimensions and 0.1 dropout. The output layer is adapted based on the loss function:
\begin{itemize}
    \item MSE baseline: Single output (predicted mean)
    \item Huber: Single output (predicted mean) with robust loss function that combines MSE for small errors and MAE for large errors, reducing sensitivity to outliers
    \item Gaussian NLL: Two outputs (mean and log-variance, transformed via softplus)
    \item Student-t NLL: Three outputs (mean, log-scale, log-degrees of freedom)
    \item Quantile Loss: Multiple outputs for different quantiles
    \item Mixture of Distributions: Outputs for mixture weights, means, and variances
    \item Adaptive Conformal Prediction: Base predictor (MSE) + calibration set for intervals
    \item Deep Evidential Regression: Four outputs for NIG parameters $(\gamma, \nu, \alpha, \beta)$
\end{itemize}

For models predicting location parameters (means), we apply bounded predictions when specified in the configuration. The location parameter is constrained using $\mu_{pred} = 3.0 \cdot \tanh(\text{head\_output})$, which bounds predicted returns to the range $[-3, 3]$ in standardized 
units (approximately $\pm3$ standard deviations). This bounded range is designed to match the normalized target data range, ensuring predictions remain within the feasible output space while allowing 
the model to capture the full range of observed returns. This also contributes to better training stability by preventing the model from attempting to predict values outside the observed data range. Other distributional parameters follow standard positivity constraints using Softplus activations where required.

\noindent\textbf{Training Configuration for Reproducibility:} Please refer to Appendix
for reproducibility.

\noindent\textbf{Trading Strategy:} We implement a directional trading strategy where positions are taken based on predicted returns:
\begin{itemize}
    \item Long (BUY): If predicted return $> 0$
    \item Short (SELL): If predicted return $\leq 0$
    \item PnL Calculation: $\text{PnL} = \text{signal} \times \text{actual\_return} \times 100$ (in basis points)
\end{itemize}

\noindent\textbf{Uncertainty-Aware Filtering:} For probabilistic methods, we evaluate an additional strategy where trades are executed only when predicted total uncertainty is below the 75th percentile threshold. This filters out high-uncertainty predictions, allowing us to assess the value of uncertainty quantification for risk-aware trading decisions.

\noindent\textbf{Evaluation Protocol:} All metrics are calculated on the test set using the best model checkpoint (selected based on validation loss). Predictions are denormalized before computing trading metrics to ensure results are interpretable in the original return scale. For trading metrics, we align the complete test set with predictions while accounting for the lookback window offset. Symbol-specific metrics are computed independently 

%% file: 3.results.tex
This section presents our evaluation and analysis for the cryptocurrency dataset (BTC). Remarkably, we also provide extensive experimental results (i.e., forecasting accuracy and trading performance) for other datasets / cryptocurrencies (ADA, BNB, DASH, DOGE, ETH, LTC, SOL, USDC, USDT, and XRP) in Appendix.

\subsection{Forecasting Accuracy Comparison}

Table~\ref{tab:accuracy_comparison} presents point prediction accuracy for BTC 
across all methods, establishing that probabilistic methods maintain 
competitive forecasting performance. Evidential Regression (ProbFM core) 
achieves comparable RMSE and MAE to baseline methods, with slightly lower 
correlation. 

The comparable performance demonstrates that our uncertainty-focused approach does not sacrifice predictive accuracy for enhanced uncertainty quantification capabilities.

\vspace{-1ex}

\begin{table}[ht]
\centering
\caption{Forecasting Accuracy Comparison for BTC}
\vspace{-2ex}
\label{tab:accuracy_comparison}
\resizebox{\linewidth}{!}{%
\begin{tabular}{lcccc}
\toprule
\textbf{Method} & \textbf{RMSE} & \textbf{MAE}  & \textbf{Correlation} \\
\midrule
MSE (Baseline) & 0.044 & 0.03 & -0.309 \\
Huber & 0.044 & 0.029 & -0.394 \\
Gaussian NLL & 0.044 & 0.029 & -0.192 \\
Student-t NLL & 0.045 & 0.03 & -0.529 \\
Quantile Loss & 0.044 & 0.029 & -0.395 \\
Adaptive Conformal (MSE) & 0.044 & 0.03 & -0.314 \\
\textbf{Evidential Regression (ProbFM core)} & 0.045 & 0.03 & -0.536 \\
\bottomrule
\end{tabular}}
\end{table}

\begin{figure}[htbp]
    \centering
    \includegraphics[width=\columnwidth]{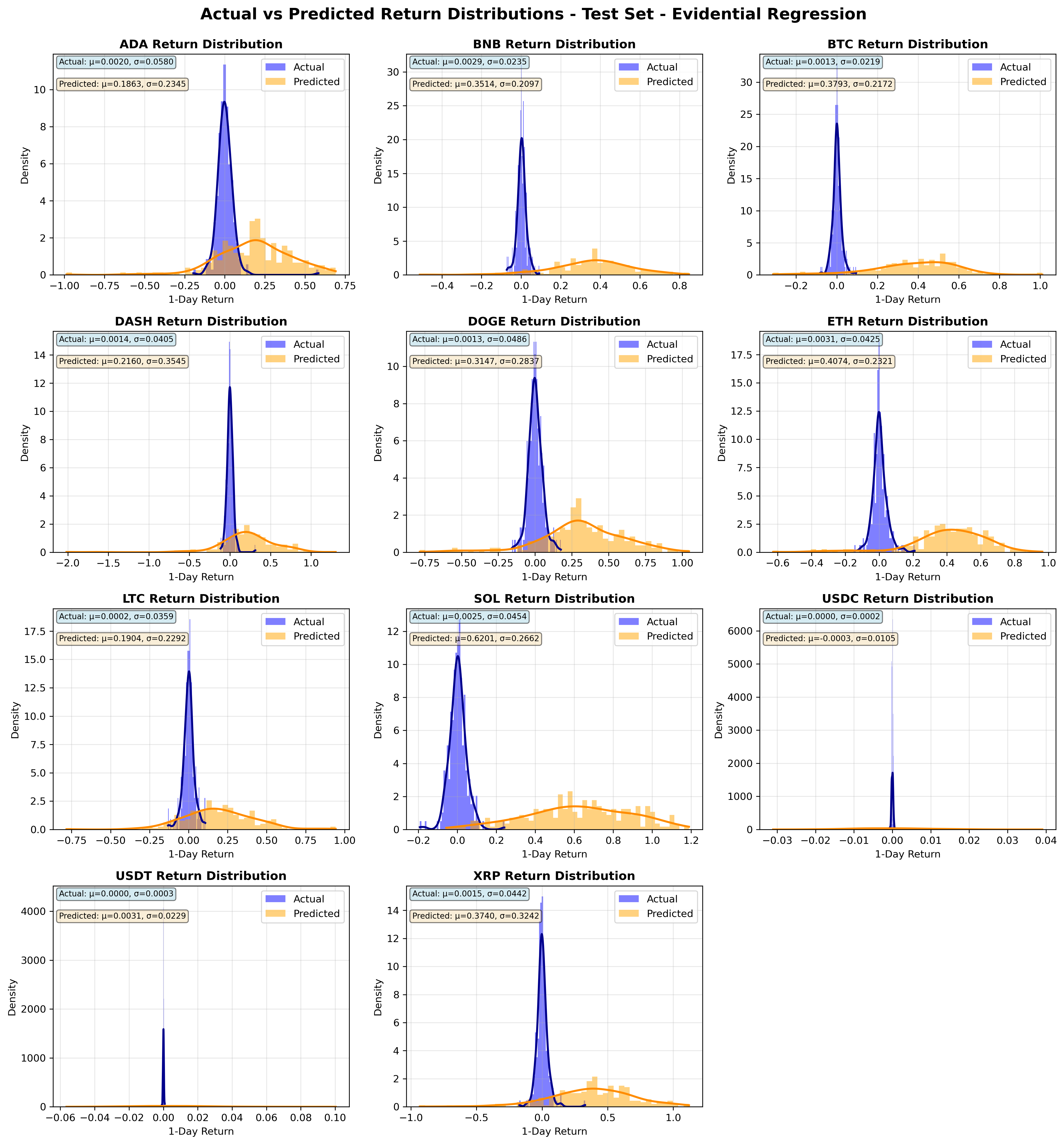}
    \caption{Distribution of actual versus predicted 1-day returns across all cryptocurrency assets using Evidential Regression. (\textmd{\textit{Best viewed in color}}).}
    \label{fig:prediction_distribution}
    \vspace{-2ex}
\end{figure}

Figure~\ref{fig:prediction_distribution} presents the distribution of actual versus predicted 1-day returns across all cryptocurrency assets using the Evidential Regression. The predicted returns represent point estimates ($\mu$) from the Evidential Regression model, while actual returns are ground truth values from the test set. Notably, predicted returns exhibit significantly wider distributions compared to actual returns across most volatile cryptocurrencies.

\vspace{-2ex}
\subsection{Uncertainty Calibration Performance}

Table~\ref{tab:calibration_comparison} presents probabilistic metrics for BTC 
across all methods, revealing key differences in uncertainty quantification 
behavior. While Evidential Regression achieves competitive CRPS (2.65) 
compared to Gaussian NLL (2.21), it exhibits distinctly different calibration 
characteristics

\begin{table}[ht]
\caption{Probabilistic Metrics Comparison for BTC}
\vspace{-2ex}
\label{tab:calibration_comparison}
\begin{center}
\resizebox{\linewidth}{!}{%
\begin{tabular}{lcccc}
\toprule
\textbf{Method} & \textbf{CRPS} & \textbf{PICP (95\%)} & \textbf{Sharpness (95\%)} & \textbf{Unc-Err Corr} \\
\midrule
MSE (Baseline) & NA & NA & NA & NA \\ 
Huber & NA & NA & NA & NA \\ 
Gaussian NLL & 2.21 & 0.95 & 15.22 & 0.12 \\ 
Student-t NLL & 3.17 & 0.95 & 15.48 & 0.11 \\ 
Quantile Loss & NA & 0.75 & 7.92 & 0.17 \\ 
Adaptive Conformal (MSE) & 2.26 & 0.81 & 9.67 & 0.09 \\ 
\textbf{Evidential Regression (ProbFM core)} & 2.65 & 0.46 & 3.9 & 0 \\ 
\bottomrule
\end{tabular}}
\end{center}
\vspace{-2ex}
\end{table}

\subsection{Trading Performance Evaluation}

Table \ref{tab:trading_performance} presents risk-adjusted trading performance for BTC across all methods. Evidential Regression demonstrates superior performance with an Annual Sharpe ratio of 1.33 and Annual Sortino ratio of 2.27, outperforming the MSE baseline (0.90 and 1.52, respectively) and other probabilistic methods. The Sortino ratio, which specifically penalizes downside volatility rather than total volatility, shows that Evidential Regression's uncertainty estimates effectively identify and avoid high-risk predictions. This is further evidenced by the competitive Calmar ratio of 3.04, indicating strong risk-adjusted returns relative to maximum drawdown. 
Evidential Regression maintains the highest win rate (0.52) alongside superior downside risk management. Notably, the maximum drawdown remains consistent across most methods (-15.14 bps), suggesting that the performance differences stem from better prediction accuracy and uncertainty-aware position sizing rather than simply avoiding extreme losses. The combination of highest Sharpe and Sortino ratios demonstrates that the explicit epistemic-aleatoric uncertainty decomposition in Evidential Regression provides actionable information for trading decisions, enabling the model to maintain exposure during confident predictions while avoiding positions during uncertain market regimes.

\begin{table}[h]
\caption{Trading Performance Comparison for BTC}
\vspace{-3ex}
\label{tab:trading_performance}
\begin{center}
\resizebox{\linewidth}{!}{%
\begin{tabular}{lccccccc}
\toprule
\textbf{Method} & \textbf{Annual Sharpe} & \textbf{Annual Sortino} & \textbf{Max DD (bps)} & \textbf{Calmar} & \textbf{Win Rate} \\
\midrule
MSE (Baseline) & 0.9 & 1.52 & -15.89 & 1.98 & 0.51 \\ 
Huber & 0.59 & 0.93 & -15.14 & 1.35 & 0.51 \\ 
Gaussian NLL & 0.49 & 0.77 & -15.14 & 1.12 & 0.5 \\ 
Student-t NLL & 0.7 & 1.08 & -15.89 & 1.41 & 0.51 \\ 
Quantile Loss & 1.13 & 1.9 & -15.14 & 2.6 & 0.52 \\ 
Adaptive Conformal (MSE) & 0.98 & 1.64 & -15.14 & 2.25 & 0.51 \\ 
\textbf{Evidential Regression (ProbFM core)} & 1.33 & 2.27 & -15.14 & 3.04 & 0.52 \\ \bottomrule
\end{tabular}}
\end{center}
\end{table}

\vspace{-3ex}

%% file: 6.conclusion.tex
We have introduced ProbFM, a transformer-based probabilistic foundation model that integrates Deep Evidential Regression (DER) with 
NIG priors for principled uncertainty quantification in time series forecasting. This work represents the first application of evidential regression to time series foundation model architectures, addressing fundamental limitations in existing approaches that either impose restrictive distributional assumptions or conflate different sources of uncertainty.

Through rigorous controlled evaluation using a consistent LSTM architecture across five probabilistic methods, we isolated the contribution of Deep Evidential Regression independent of architectural complexity. Our comprehensive preprocessing pipeline—including per-symbol outlier truncation and standardization with bounded predictions—ensures stable training while maintaining model expressiveness. Evaluation on cryptocurrency return forecasting demonstrates that Deep Evidential Regression-based models achieve competitive point prediction accuracy while providing superior  risk-adjusted trading performance.

We also outline the related work and limitations of our work and propose avenues for future research in Appendix. 

%% file: 7.disclaimer.tex
JPMorganChase makes no representation and warranty whatsoever and disclaims all liability, for the completeness, accuracy or reliability of the information contained herein. Any views or opinions expressed herein are solely those of the authors. This document is not intended as investment research or investment advice, or a recommendation, offer or solicitation for the purchase or sale of any security, financial instrument, financial product or service, or to be used in any way for evaluating the merits of participating in any transaction, and shall not constitute a solicitation under any jurisdiction or to any person, if such solicitation under such jurisdiction or to such person would be unlawful.

%% file: 8.appendix.tex
\section{ProbFM Architecture Pipeline}\label{appendix:pipeline}

This appendix provides the ProbFM architecture pipeline, illustrating the end-to-end methodology from input processing to single-pass inference.

\begin{figure}[H]
\centering
\includegraphics[width=0.47\textwidth,height=0.52\textheight,keepaspectratio]
{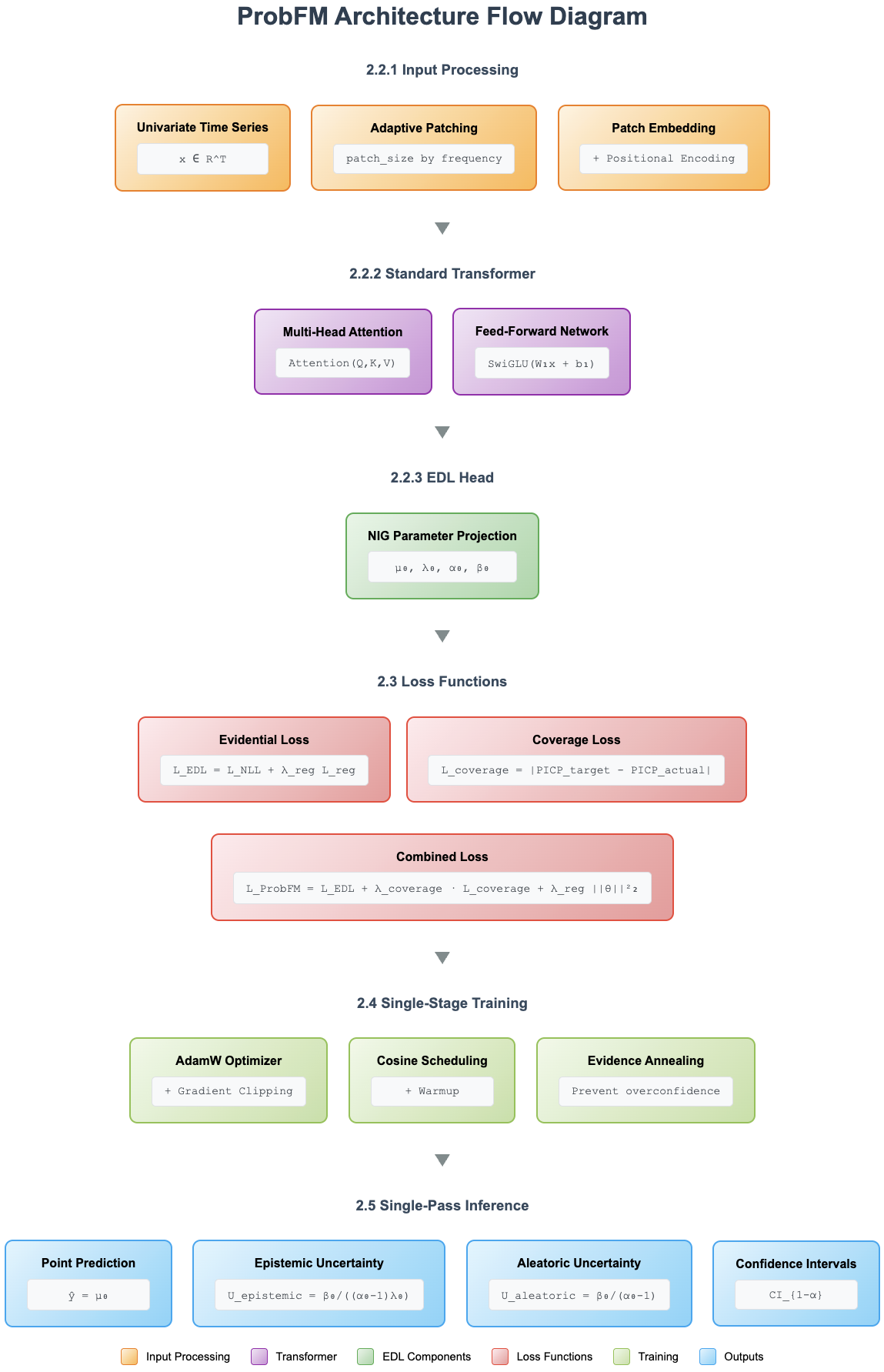}
\caption{Overview of the ProbFM architecture pipeline, illustrating the end-to-end methodology from input processing to single-pass inference
(\textmd{\textit{Best viewed in color}}).}
\label{fig:probfm_architecture}
\end{figure}

\section{Evaluation Metrics and Protocols}\label{appendix:metrics}

\subsection{Accuracy Metrics}\label{appendix:metrics_accuracy}
\begin{itemize}
\item \textbf{Root Mean Squared Error (RMSE)}: Square root of MSE, interpretable in the same units as returns
\item \textbf{Mean Absolute Error (MAE)}: Captures the average absolute prediction error
\item \textbf{Pearson Correlation}: Linear correlation between predicted and actual returns
\end{itemize}

\subsection{Probabilistic Metrics}\label{appendix:metrics_prob}

\begin{itemize}
\item \textbf{Continuous Ranked Probability Score (CRPS)}: Measures the distance between predicted and actual distributions. Lower values indicate better probabilistic forecasts. For Gaussian predictions, CRPS is calculated as:

  \begin{equation}
  \text{CRPS} = \sigma \left( z(2\Phi(z) - 1) + 2\phi(z) - \frac{1}{\sqrt{\pi}} \right)
  \end{equation}
  
  where $z = \frac{y - \mu}{\sigma}$, $\mu$ is the predicted mean, $\sigma$ is the predicted standard deviation, and $\Phi$ and $\phi$ are the CDF and PDF of the standard normal distribution.

\item \textbf{Prediction Interval Coverage Probability (PICP)}: The proportion of actual values falling within predicted confidence intervals. For well-calibrated models, PICP should equal the target confidence level (e.g., 95\% for a 95\% confidence interval).

\item \textbf{Sharpness}: Average width of prediction intervals, measuring prediction precision. Lower sharpness indicates more confident predictions.

\item \textbf{Uncertainty-Error Correlation}: Pearson correlation between predicted uncertainty (standard deviation) and actual prediction errors. Positive correlation indicates the model appropriately assigns higher uncertainty to harder-to-predict samples.

\end{itemize}

\subsection{Trading Metrics}\label{appendix:metrics_trade}

\begin{itemize}
\item \textbf{Sharpe Ratio}: Risk-adjusted return calculated as $\frac{\mu_r}{\sigma_r}$ where $\mu_r$ is mean return and $\sigma_r$ is return standard deviation. We report both daily and annualized (multiplied by $\sqrt{252}$) versions.

\item \textbf{Sortino Ratio}: Downside risk-adjusted return calculated as $\frac{\mu_r}{\sigma_{\text{downside}}}$ where $\mu_r$ is mean return and $\sigma_{\text{downside}}$ is the standard deviation of negative returns only. Unlike Sharpe ratio which penalizes all volatility, Sortino ratio only penalizes downside volatility. We report annualized versions (multiplied by $\sqrt{252}$).

\item \textbf{Maximum Drawdown}: Largest peak-to-trough decline in cumulative PnL, measuring downside risk exposure.

\item \textbf{Calmar Ratio}: Ratio of annualized return to absolute maximum drawdown, providing a risk-adjusted performance measure that specifically accounts for tail risk.

\item \textbf{Win Rate}: Proportion of profitable trades, indicating prediction directional accuracy.

\end{itemize}

\section{Training Configuration for Reproducibility}\label{appendix:reproduce}

In this section, we present the hyperparameters chosen after comprehensive tuning to support reproducibility. All models are trained with the following configuration:

\subsection{Model Architecture}
\begin{itemize}
    \item \textbf{Base Architecture:} 1-layer LSTM
    \item \textbf{Hidden Dimensions:} 32
    \item \textbf{Dropout Rate:} 0.1
    \item \textbf{Lookback Window:} 50 time steps
    \item \textbf{Output Dimension:} Varies by loss function (1 for MSE/Huber, 2 for Gaussian NLL, 3 for Student-t NLL, 4 for Evidential Regression)
\end{itemize}

\subsection{Optimization Hyperparameters}
\begin{itemize}
    \item \textbf{Optimizer:} Adam
    \item \textbf{Learning Rate:} 0.001
    \item \textbf{Batch Size:} 128
    \item \textbf{Maximum Epochs:} 50
    \item \textbf{Early Stopping Patience:} 10 epochs
    \item \textbf{Gradient Clipping:} Maximum norm 1.0
    \item \textbf{Weight Decay:} 0.001 (L2 regularization)
\end{itemize}

\subsection{Data Processing}\label{appendix:data_preprocessing}
\begin{itemize}
    \item \textbf{Temporal Split:} Data up to 2023 for train/validation, 2024+ for test
    \item \textbf{Validation Split:} 20\% of training data (per symbol, maintaining temporal order)
    \item \textbf{Outlier Truncation:} 1st and 99th percentiles (per symbol, calculated from training data)
    \item \textbf{Standardization:} Zero mean, unit variance (per symbol, using training statistics)
\end{itemize}

\subsection{Method-Specific Hyperparameters}
\begin{itemize}
    \item \textbf{Evidential Regression:}
    \begin{itemize}
        \item Regularization weight ($\lambda_{evd}$): 0.1
        \item Bounded predictions: $\gamma = 3.0 \cdot \tanh(\text{head\_output})$
        \item Parameter constraints: $\nu = \text{Softplus}(\cdot) + 0.01$, $\alpha = \text{Softplus}(\cdot) + 1.0$, $\beta = \text{Softplus}(\cdot) + 0.01$
    \end{itemize}
    \item \textbf{Huber Loss:}
    \begin{itemize}
        \item Delta parameter ($\delta$): 1.0 (default PyTorch value)
    \end{itemize}
    \item \textbf{Uncertainty-Aware Trading:}
    \begin{itemize}
        \item Uncertainty threshold: 75th percentile of total uncertainty
        \item Hold threshold: 0.0 (directional trading without neutral zone)
    \end{itemize}
\end{itemize}

\subsection{Implementation Details}
\begin{itemize}
    \item \textbf{Random Seed:} 0 (for reproducibility)
    \item \textbf{Framework:} PyTorch
    \item \textbf{Number of Workers:} 0 (DataLoader)
    \item \textbf{Model Selection:} Best validation loss
\end{itemize}

\textbf{Note:} While weight decay and dropout were both considered during hyperparameter tuning, the final configuration uses weight decay 0.001 and dropout 0.1, as this combination yielded the best empirical results across all methods. The consistent architecture across all loss functions ensures that performance differences can be attributed to the uncertainty quantification strategy rather than architectural variations.

\section{Forecasting Accuracy and Trading Performance}\label{appendix:forecasting}

This appendix provides extensive experiments conducted on other cryptocurrencies besides BTC.

\begin{table}[H]
\centering
\caption{Forecasting Accuracy Comparison for ADA}
\vspace{-2ex}
\label{tab:accuracy_comparison_ada}
\resizebox{\linewidth}{!}{%
\begin{tabular}{lcccc}
\toprule
\textbf{Method} & \textbf{RMSE} & \textbf{MAE}  & \textbf{Correlation} \\
\midrule
MSE (Baseline) & 0.05 & 0.037 & -0.488 \\
Huber & 0.049 & 0.037 & -0.524 \\
Gaussian NLL & 0.049 & 0.036 & -0.361 \\
Student-t NLL & 0.05 & 0.037 & -0.614 \\
Quantile Loss & 0.049 & 0.036 & -0.507 \\
Adaptive Conformal (MSE) & 0.049 & 0.037 & -0.491 \\
\textbf{Evidential Regression (ProbFM core)} & 0.05 & 0.037 & -0.642 \\
\bottomrule
\end{tabular}}
\end{table}

\begin{table}[H]
\centering
\caption{Forecasting Accuracy Comparison for BNB}
\vspace{-2ex}
\label{tab:accuracy_comparison_bnb}
\resizebox{\linewidth}{!}{%
\begin{tabular}{lcccc}
\toprule
\textbf{Method} & \textbf{RMSE} & \textbf{MAE}  & \textbf{Correlation} \\
\midrule
MSE (Baseline) & 0.022 & 0.016 & 0.019 \\
Huber & 0.022 & 0.015 & 0.091 \\
Gaussian NLL & 0.022 & 0.016 & -0.028 \\
Student-t NLL & 0.022 & 0.015 & 0.086 \\
Quantile Loss & 0.022 & 0.015 & 0.057 \\
Adaptive Conformal (MSE) & 0.022 & 0.016 & 0.019 \\
\textbf{Evidential Regression (ProbFM core)} & 0.022 & 0.016 & 0.052 \\
\bottomrule
\end{tabular}}
\end{table}

\vspace{-4ex}
\begin{table}[H]
\centering
\caption{Forecasting Accuracy Comparison for DASH}
\vspace{-2ex}
\label{tab:accuracy_comparison_dash}
\resizebox{\linewidth}{!}{%
\begin{tabular}{lcccc}
\toprule
\textbf{Method} & \textbf{RMSE} & \textbf{MAE}  & \textbf{Correlation} \\
\midrule
MSE (Baseline) & 0.025 & 0.018 & -0.348 \\
Huber & 0.025 & 0.018 & -0.411 \\
Gaussian NLL & 0.024 & 0.018 & -0.233 \\
Student-t NLL & 0.026 & 0.019 & -0.459 \\
Quantile Loss & 0.024 & 0.018 & -0.411 \\
Adaptive Conformal (MSE) & 0.025 & 0.018 & -0.349 \\
\textbf{Evidential Regression (ProbFM core)} & 0.025 & 0.018 & -0.454 \\
\bottomrule
\end{tabular}}
\end{table}

\vspace{-4ex}
\begin{table}[H]
\centering
\caption{Forecasting Accuracy Comparison for DOGE}
\vspace{-2ex}
\label{tab:accuracy_comparison_doge}
\resizebox{\linewidth}{!}{%
\begin{tabular}{lcccc}
\toprule
\textbf{Method} & \textbf{RMSE} & \textbf{MAE}  & \textbf{Correlation} \\
\midrule
MSE (Baseline) & 0.004 & 0.003 & 0.034 \\
Huber & 0.002 & 0.002 & 0.051 \\
Gaussian NLL & 0.003 & 0.002 & 0.021 \\
Student-t NLL & 0.003 & 0.002 & 0.055 \\
Quantile Loss & 0.001 & 0.001 & 0.046 \\
Adaptive Conformal (MSE) & 0.004 & 0.004 & 0.034 \\
\textbf{Evidential Regression (ProbFM core)} & 0.004 & 0.004 & 0.031 \\
\bottomrule
\end{tabular}}
\end{table}

\vspace{-4ex}
\begin{table}[H]
\centering
\caption{Forecasting Accuracy Comparison for ETH}
\vspace{-2ex}
\label{tab:accuracy_comparison_eth}
\resizebox{\linewidth}{!}{%
\begin{tabular}{lcccc}
\toprule
\textbf{Method} & \textbf{RMSE} & \textbf{MAE}  & \textbf{Correlation} \\
\midrule
MSE (Baseline) & 0.005 & 0.004 & 0.025 \\
Huber & 0.003 & 0.003 & 0.033 \\
Gaussian NLL & 0.005 & 0.004 & 0.014 \\
Student-t NLL & 0.003 & 0.003 & 0.031 \\
Quantile Loss & 0.001 & 0.001 & 0.031 \\
Adaptive Conformal (MSE) & 0.005 & 0.005 & 0.025 \\
\textbf{Evidential Regression (ProbFM core)} & 0.005 & 0.004 & 0.017 \\
\bottomrule
\end{tabular}}
\end{table}

\vspace{-4ex}
\begin{table}[H]
\centering
\caption{Forecasting Accuracy Comparison for LTC}
\vspace{-2ex}
\label{tab:accuracy_comparison_ltc}
\resizebox{\linewidth}{!}{%
\begin{tabular}{lcccc}
\toprule
\textbf{Method} & \textbf{RMSE} & \textbf{MAE}  & \textbf{Correlation} \\
\midrule
MSE (Baseline) & 0.043 & 0.03 & -0.377 \\
Huber & 0.043 & 0.03 & -0.51 \\
Gaussian NLL & 0.042 & 0.03 & -0.21 \\
Student-t NLL & 0.043 & 0.03 & -0.514 \\
Quantile Loss & 0.042 & 0.029 & -0.481 \\
Adaptive Conformal (MSE) & 0.043 & 0.03 & -0.378 \\
\textbf{Evidential Regression (ProbFM core)} & 0.043 & 0.03 & -0.5 \\
\bottomrule
\end{tabular}}
\end{table}

\vspace{-4ex}
\begin{table}[H]
\centering
\caption{Forecasting Accuracy Comparison for SOL}
\vspace{-2ex}
\label{tab:accuracy_comparison_sol}
\resizebox{\linewidth}{!}{%
\begin{tabular}{lcccc}
\toprule
\textbf{Method} & \textbf{RMSE} & \textbf{MAE}  & \textbf{Correlation} \\
\midrule
MSE (Baseline) & 0.044 & 0.033 & -0.029 \\
Huber & 0.043 & 0.033 & -0.012 \\
Gaussian NLL & 0.044 & 0.033 & -0.051 \\
Student-t NLL & 0.043 & 0.033 & 0.027 \\
Quantile Loss & 0.043 & 0.033 & -0.02 \\
Adaptive Conformal (MSE) & 0.044 & 0.033 & -0.027 \\
\textbf{Evidential Regression (ProbFM core)} & 0.044 & 0.033 & 0.035 \\
\bottomrule
\end{tabular}}
\end{table}

\vspace{-4ex}
\begin{table}[H]
\centering
\caption{Forecasting Accuracy Comparison for USDC}
\vspace{-2ex}
\label{tab:accuracy_comparison_usdc}
\resizebox{\linewidth}{!}{%
\begin{tabular}{lcccc}
\toprule
\textbf{Method} & \textbf{RMSE} & \textbf{MAE}  & \textbf{Correlation} \\
\midrule
MSE (Baseline) & 0.036 & 0.025 & 0.022 \\
Huber & 0.036 & 0.025 & 0.027 \\
Gaussian NLL & 0.036 & 0.025 & -0.032 \\
Student-t NLL & 0.036 & 0.025 & 0.02 \\
Quantile Loss & 0.036 & 0.025 & 0.027 \\
Adaptive Conformal (MSE) & 0.036 & 0.025 & 0.025 \\
\textbf{Evidential Regression (ProbFM core)} & 0.036 & 0.025 & 0.026 \\
\bottomrule
\end{tabular}}
\end{table}

\vspace{-4ex}
\begin{table}[H]
\centering
\caption{Forecasting Accuracy Comparison for USDT}
\vspace{-2ex}
\label{tab:accuracy_comparison_usdt}
\resizebox{\linewidth}{!}{%
\begin{tabular}{lcccc}
\toprule
\textbf{Method} & \textbf{RMSE} & \textbf{MAE}  & \textbf{Correlation} \\
\midrule
MSE (Baseline) & 0.039 & 0.028 & 0.052 \\
Huber & 0.039 & 0.028 & 0.06 \\
Gaussian NLL & 0.039 & 0.028 & 0.027 \\
Student-t NLL & 0.039 & 0.028 & 0.048 \\
Quantile Loss & 0.039 & 0.028 & 0.058 \\
Adaptive Conformal (MSE) & 0.039 & 0.028 & 0.052 \\
\textbf{Evidential Regression (ProbFM core)} & 0.039 & 0.028 & 0.051 \\
\bottomrule
\end{tabular}}
\end{table}

\vspace{-4ex}
\begin{table}[H]
\centering
\caption{Forecasting Accuracy Comparison for XRP}
\vspace{-2ex}
\label{tab:accuracy_comparison_xrp}
\resizebox{\linewidth}{!}{%
\begin{tabular}{lcccc}
\toprule
\textbf{Method} & \textbf{RMSE} & \textbf{MAE}  & \textbf{Correlation} \\
\midrule
MSE (Baseline) & 0.044 & 0.032 & 0.106 \\
Huber & 0.044 & 0.032 & 0.105 \\
Gaussian NLL & 0.044 & 0.032 & 0.081 \\
Student-t NLL & 0.044 & 0.032 & 0.065 \\
Quantile Loss & 0.044 & 0.032 & 0.056 \\
Adaptive Conformal (MSE) & 0.044 & 0.032 & 0.104 \\
\textbf{Evidential Regression (ProbFM core)} & 0.044 & 0.032 & 0.054 \\
\bottomrule
\end{tabular}}
\end{table}

\vspace{-4ex}
\begin{table}[H]
\caption{Probabilistic Metrics Comparison for ADA}
\vspace{-2ex}
\label{tab:calibration_comparison_ada}
\begin{center}
\resizebox{\linewidth}{!}{%
\begin{tabular}{lcccc}
\toprule
\textbf{Method} & \textbf{CRPS} & \textbf{PICP (95\%)} & \textbf{Sharpness (95\%)} & \textbf{Unc-Err Corr} \\
\midrule
MSE (Baseline) & NA & NA & NA & NA \\ 
Huber & NA & NA & NA & NA \\ 
Gaussian NLL & 2.67 & 0.92 & 17.1 & 0.13 \\ 
Student-t NLL & 3.81 & 0.89 & 16.45 & 0.19 \\ 
Quantile Loss & NA & 0.6814 & 8.9071 & 0.07 \\ 
Adaptive Conformal (MSE) & 2.8 & 0.75 & 10.66 & 0.19 \\ 
\textbf{Evidential Regression (ProbFM core)} & 3.39 & 0.31 & 3.32 & 0.09 \\ 
\bottomrule
\end{tabular}}
\end{center}
\end{table}

\vspace{-4ex}
\begin{table}[H]
\caption{Probabilistic Metrics Comparison for BNB}
\vspace{-2ex}
\label{tab:calibration_comparison_bnb}
\begin{center}
\resizebox{\linewidth}{!}{%
\begin{tabular}{lcccc}
\toprule
\textbf{Method} & \textbf{CRPS} & \textbf{PICP (95\%)} & \textbf{Sharpness (95\%)} & \textbf{Unc-Err Corr} \\
\midrule
MSE (Baseline) & NA & NA & NA & NA \\ 
Huber & NA & NA & NA & NA \\ 
Gaussian NLL & 1.41 & 0.99 & 16.63 & -0.05 \\ 
Student-t NLL & 2.17 & 0.98 & 16.11 & -0.02 \\ 
Quantile Loss & NA & 0.93 & 8.67 & -0.04 \\ 
Adaptive Conformal (MSE) & 1.24 & 0.95 & 10.58 & -0.09 \\ 
\textbf{Evidential Regression (ProbFM core)} & 1.3 & 0.68 & 3.63 & -0.12 \\ 
\bottomrule
\end{tabular}}
\end{center}
\end{table}

\vspace{-4ex}
\begin{table}[H]
\caption{Probabilistic Metrics Comparison for DASH}
\vspace{-2ex}
\label{tab:calibration_comparison_dash}
\begin{center}
\resizebox{\linewidth}{!}{%
\begin{tabular}{lcccc}
\toprule
\textbf{Method} & \textbf{CRPS} & \textbf{PICP (95\%)} & \textbf{Sharpness (95\%)} & \textbf{Unc-Err Corr} \\
\midrule
MSE (Baseline) & NA & NA & NA & NA \\ 
Huber & NA & NA & NA & NA \\ 
Gaussian NLL & 1.66 & 1 & 21.23 & 0.19 \\ 
Student-t NLL & 2.77 & 1 & 21.33 & 0.29 \\ 
Quantile Loss & NA & 0.96 & 11.07 & 0.21 \\ 
Adaptive Conformal (MSE) & 1.39 & 0.98 & 13.26 & 0.25 \\ 
\textbf{Evidential Regression (ProbFM core)} & 1.57 & 0.59 & 3.49 & 0.04 \\ 
\bottomrule
\end{tabular}}
\end{center}
\end{table}

\vspace{-4ex}
\begin{table}[H]
\caption{Probabilistic Metrics Comparison for DOGE}
\vspace{-2ex}
\label{tab:calibration_comparison_doge}
\begin{center}
\resizebox{\linewidth}{!}{%
\begin{tabular}{lcccc}
\toprule
\textbf{Method} & \textbf{CRPS} & \textbf{PICP (95\%)} & \textbf{Sharpness (95\%)} & \textbf{Unc-Err Corr} \\
\midrule
MSE (Baseline) & NA & NA & NA & NA \\ 
Huber & NA & NA & NA & NA \\ 
Gaussian NLL & 1.23 & 1 & 20.59 & 0.7 \\ 
Student-t NLL & 2.12 & 1 & 21.38 & 0.06 \\ 
Quantile Loss & NA & 1 & 10.76 & 0.12 \\ 
Adaptive Conformal (MSE) & 0.78 & 1 & 12.71 & 0.29 \\ 
\textbf{Evidential Regression (ProbFM core)} & 0.25 & 1 & 3.77 & 0.18 \\ 
\bottomrule
\end{tabular}}
\end{center}
\end{table}

\vspace{-4ex}
\begin{table}[H]
\caption{Probabilistic Metrics Comparison for ETH}
\vspace{-2ex}
\label{tab:calibration_comparison_eth}
\begin{center}
\resizebox{\linewidth}{!}{%
\begin{tabular}{lcccc}
\toprule
\textbf{Method} & \textbf{CRPS} & \textbf{PICP (95\%)} & \textbf{Sharpness (95\%)} & \textbf{Unc-Err Corr} \\
\midrule
MSE (Baseline) & NA & NA & NA & NA \\ 
Huber & NA & NA & NA & NA \\ 
Gaussian NLL & 1.17 & 1 & 19.3 & 0.85 \\ 
Student-t NLL & 1.96 & 1 & 20.46 & 0.03 \\ 
Quantile Loss & NA & 1 & 10.16 & 0.54 \\ 
Adaptive Conformal (MSE) & 0.73 & 1 & 11.72 & 0.49 \\ 
\textbf{Evidential Regression (ProbFM core)} & 0.29 & 1 & 4.17 & 0.22 \\ 
\bottomrule
\end{tabular}}
\end{center}
\end{table}

\vspace{-4ex}
\begin{table}[H]
\caption{Probabilistic Metrics Comparison for LTC}
\vspace{-2ex}
\label{tab:calibration_comparison_ltc}
\begin{center}
\resizebox{\linewidth}{!}{%
\begin{tabular}{lcccc}
\toprule
\textbf{Method} & \textbf{CRPS} & \textbf{PICP (95\%)} & \textbf{Sharpness (95\%)} & \textbf{Unc-Err Corr} \\
\midrule
MSE (Baseline) & NA & NA & NA & NA \\ 
Huber & NA & NA & NA & NA \\ 
Gaussian NLL & 2.27 & 0.95 & 17.53 & 0.08 \\ 
Student-t NLL & 3.31 & 0.93 & 17.02 & 0.07 \\ 
Quantile Loss & NA & 0.78 & 9.19 & 0.03 \\ 
Adaptive Conformal (MSE) & 2.27 & 0.85 & 10.84 & 0.15 \\ 
\textbf{Evidential Regression (ProbFM core)} & 2.69 & 0.46 & 3.56 & 0.03 \\ 
\bottomrule
\end{tabular}}
\end{center}
\end{table}

\vspace{-4ex}
\begin{table}[H]
\caption{Probabilistic Metrics Comparison for SOL}
\vspace{-2ex}
\label{tab:calibration_comparison_sol}
\begin{center}
\resizebox{\linewidth}{!}{%
\begin{tabular}{lcccc}
\toprule
\textbf{Method} & \textbf{CRPS} & \textbf{PICP (95\%)} & \textbf{Sharpness (95\%)} & \textbf{Unc-Err Corr} \\
\midrule
MSE (Baseline) & NA & NA & NA & NA \\ 
Huber & NA & NA & NA & NA \\ 
Gaussian NLL & 2.45 & 0.97 & 21.87 & 0.1 \\ 
Student-t NLL & 3.71 & 0.98 & 22.47 & 0.04 \\ 
Quantile Loss & NA & 0.85 & 11.27 & 0.16 \\ 
Adaptive Conformal (MSE) & 2.43 & 0.88 & 13.96 & 0.01 \\ 
\textbf{Evidential Regression (ProbFM core)} & 2.96 & 0.39 & 3.85 & -0.14 \\ 
\bottomrule
\end{tabular}}
\end{center}
\end{table}

\vspace{-4ex}
\begin{table}[H]
\caption{Probabilistic Metrics Comparison for USDC}
\vspace{-2ex}
\label{tab:calibration_comparison_usdc}
\begin{center}
\resizebox{\linewidth}{!}{%
\begin{tabular}{lcccc}
\toprule
\textbf{Method} & \textbf{CRPS} & \textbf{PICP (95\%)} & \textbf{Sharpness (95\%)} & \textbf{Unc-Err Corr} \\
\midrule
MSE (Baseline) & NA & NA & NA & NA \\ 
Huber & NA & NA & NA & NA \\ 
Gaussian NLL & 2.48 & 0.08 & 0.46 & 0.11 \\ 
Student-t NLL & 2.6 & 0.05 & 0.32 & -0.03 \\ 
Quantile Loss & NA & 0.04 & 0.23 & 0.1 \\ 
Adaptive Conformal (MSE) & 2.5 & 0.05 & 0.28 & 0.04 \\ 
\textbf{Evidential Regression (ProbFM core)} & 2.32 & 0.78 & 7.44 & -0.04 \\ 
\bottomrule
\end{tabular}}
\end{center}
\end{table}

\vspace{-4ex}
\begin{table}[H]
\caption{Probabilistic Metrics Comparison for USDT}
\vspace{-2ex}
\label{tab:calibration_comparison_usdt}
\begin{center}
\resizebox{\linewidth}{!}{%
\begin{tabular}{lcccc}
\toprule
\textbf{Method} & \textbf{CRPS} & \textbf{PICP (95\%)} & \textbf{Sharpness (95\%)} & \textbf{Unc-Err Corr} \\
\midrule
MSE (Baseline) & NA & NA & NA & NA \\ 
Huber & NA & NA & NA & NA \\ 
Gaussian NLL & 2.62 & 0.15 & 1.1 & -0.06 \\ 
Student-t NLL & 2.8 & 0.09 & 0.74 & -0.03 \\ 
Quantile Loss & NA & 0.06 & 0.57 & -0.05 \\ 
Adaptive Conformal (MSE) & 2.67 & 0.08 & 0.7 & -0.12 \\ 
\textbf{Evidential Regression (ProbFM core)} & 2.54 & 0.45 & 3.56 & -0.01 \\ 
\bottomrule
\end{tabular}}
\end{center}
\end{table}

\vspace{-4ex}
\begin{table}[H]
\caption{Probabilistic Metrics Comparison for XRP}
\vspace{-2ex}
\label{tab:calibration_comparison_xrp}
\begin{center}
\resizebox{\linewidth}{!}{%
\begin{tabular}{lcccc}
\toprule
\textbf{Method} & \textbf{CRPS} & \textbf{PICP (95\%)} & \textbf{Sharpness (95\%)} & \textbf{Unc-Err Corr} \\
\midrule
MSE (Baseline) & NA & NA & NA & NA \\ 
Huber & NA & NA & NA & NA \\ 
Gaussian NLL & 2.47 & 0.99 & 22.89 & 0.01 \\ 
Student-t NLL & 3.7 & 0.96 & 22.06 & 0 \\ 
Quantile Loss & NA & 0.85 & 12.17 & 0 \\ 
Adaptive Conformal (MSE) & 2.39 & 0.89 & 13.68 & -0.04 \\ 
\textbf{Evidential Regression (ProbFM core)} & 2.9 & 0.36 & 3.25 & -0.03 \\ 
\bottomrule
\end{tabular}}
\end{center}
\end{table}

\vspace{-4ex}
\begin{table}[H]
\caption{Trading Performance Comparison for ADA}
\vspace{-2ex}
\label{tab:trading_performance_ada}
\begin{center}
\resizebox{\linewidth}{!}{%
\begin{tabular}{lccccccc}
\toprule
\textbf{Method} & \textbf{Annual Sharpe} & \textbf{Annual Sortino} & \textbf{Max DD (bps)} & \textbf{Calmar} & \textbf{Win Rate} \\
\midrule
MSE (Baseline) & 1.02 & 2.34 & -49.9 & 1.88 & 0.51 \\
Huber & 1.21 & 2.87 & -64.58 & 1.73 & 0.53 \\
Gaussian NLL & -0.39 & -0.4 & -60.97 & -0.6 & 0.49 \\
Student-t NLL & -0.92 & -0.96 & -9875.19 & -0.85 & 0.46 \\
Quantile Loss & -0.31 & -0.3 & -65.66 & -0.43 & 0.53 \\
Adaptive Conformal (MSE) & 0.7 & 1.59 & -48.46 & 1.33 & 0.49 \\
\textbf{Evidential Regression (ProbFM core)} & -0.83 & -0.86 & -88.94 & -0.86 & 0.46 \\ 
\bottomrule
\end{tabular}}
\end{center}
\end{table}

\vspace{-4ex}
\begin{table}[H]
\caption{Trading Performance Comparison for BNB}
\vspace{-2ex}
\label{tab:trading_performance_bnb}
\begin{center}
\resizebox{\linewidth}{!}{%
\begin{tabular}{lccccccc}
\toprule
\textbf{Method} & \textbf{Annual Sharpe} & \textbf{Annual Sortino} & \textbf{Max DD (bps)} & \textbf{Calmar} & \textbf{Win Rate} \\
\midrule
MSE (Baseline) & 2.41 & 3.35 & -21.26 & 4.22 & 0.58 \\ 
Huber & 1.49 & 1.95 & -21.26 & 2.62 & 0.56 \\ 
Gaussian NLL & 1.77 & 2.53 & -17.3 & 3.82 & 0.58 \\ 
Student-t NLL & 1.31 & 1.77 & -2131.56 & 2.31 & 0.55 \\ 
Quantile Loss & 1.09 & 1.43 & -26.64 & 1.53 & 0.55 \\ 
Adaptive Conformal (MSE) & 2.13 & 3.02 & -21.26 & 3.73 & 0.57 \\ 
\textbf{Evidential Regression (ProbFM core)} & 1.96 & 2.8 & -21.26 & 3.44 & 0.57 \\ \bottomrule
\end{tabular}}
\end{center}
\end{table}

\vspace{-4ex}
\begin{table}[H]
\caption{Trading Performance Comparison for DASH}
\vspace{-2ex}
\label{tab:trading_performance_dash}
\begin{center}
\resizebox{\linewidth}{!}{%
\begin{tabular}{lccccccc}
\toprule
\textbf{Method} & \textbf{Annual Sharpe} & \textbf{Annual Sortino} & \textbf{Max DD (bps)} & \textbf{Calmar} & \textbf{Win Rate} \\
\midrule
MSE (Baseline) & 0.49 & 0.56 & -52.36 & 0.6 & 0.52 \\ 
Huber & 1.02 & 1.15 & -44.87 & 1.46 & 0.53 \\ 
Gaussian NLL & -0.39 & -0.46 & -68.58 & -0.37 & 0.5 \\ 
Student-t NLL & -0.44 & -0.53 & -6962.52 & -0.41 & 0.48 \\ 
Quantile Loss & -0.57 & -0.67 & -73.15 & -0.5 & 0.48 \\ 
Adaptive Conformal (MSE) & -0.01 & -0.02 & -52.36 & -0.02 & 0.52 \\ 
\textbf{Evidential Regression (ProbFM core)} & -0.49 & -0.55 & -66.52 & -0.47 & 0.5 \\ \bottomrule
\end{tabular}}
\end{center}
\end{table}

\vspace{-4ex}
\begin{table}[H]
\caption{Trading Performance Comparison for DOGE}
\vspace{-2ex}
\label{tab:trading_performance_doge}
\begin{center}
\resizebox{\linewidth}{!}{%
\begin{tabular}{lccccccc}
\toprule
\textbf{Method} & \textbf{Annual Sharpe} & \textbf{Annual Sortino} & \textbf{Max DD (bps)} & \textbf{Calmar} & \textbf{Win Rate} \\
\midrule
MSE (Baseline) & -0.9 & -1.36 & -80.36 & -0.86 & 0.46 \\ 
Huber & -1.63 & -2.23 & -117.63 & -1.06 & 0.48 \\ 
Gaussian NLL & -0.35 & -0.5 & -88.63 & -0.31 & 0.5 \\ 
Student-t NLL & -0.99 & -1.33 & -10068.88 & -0.76 & 0.48 \\ 
Quantile Loss & -0.19 & -0.25 & -91.29 & -0.16 & 0.53 \\ 
Adaptive Conformal (MSE) & -0.96 & -1.39 & -83.42 & -0.88 & 0.46 \\ 
\textbf{Evidential Regression (ProbFM core)} & -0.08 & -0.12 & -63.42 & -0.1 & 0.49 \\ \bottomrule
\end{tabular}}
\end{center}
\end{table}

\vspace{-4ex}
\begin{table}[H]
\caption{Trading Performance Comparison for ETH}
\vspace{-2ex}
\label{tab:trading_performance_eth}
\begin{center}
\resizebox{\linewidth}{!}{%
\begin{tabular}{lccccccc}
\toprule
\textbf{Method} & \textbf{Annual Sharpe} & \textbf{Annual Sortino} & \textbf{Max DD (bps)} & \textbf{Calmar} & \textbf{Win Rate} \\
\midrule
MSE (Baseline) & 0.41 & 0.64 & -58.26 & 0.48 & 0.48 \\ 
Huber & 0.65 & 1 & -58.26 & 0.75 & 0.49 \\ 
Gaussian NLL & 0.64 & 1.01 & -58.26 & 0.75 & 0.48 \\ 
Student-t NLL & 1.86 & 3.33 & -2707.71 & 4.63 & 0.51 \\ 
Quantile Loss & 0.41 & 0.64 & -58.26 & 0.47 & 0.48 \\ 
Adaptive Conformal (MSE) & 0.48 & 0.75 & -58.26 & 0.56 & 0.48 \\ 
\textbf{Evidential Regression (ProbFM core)} & 1.1 & 1.84 & -38.83 & 1.91 & 0.5 \\ 
\bottomrule
\end{tabular}}
\end{center}
\end{table}

\vspace{-4ex}
\begin{table}[H]
\caption{Trading Performance Comparison for LTC}
\vspace{-2ex}
\label{tab:trading_performance_ltc}
\begin{center}
\resizebox{\linewidth}{!}{%
\begin{tabular}{lccccccc}
\toprule
\textbf{Method} & \textbf{Annual Sharpe} & \textbf{Annual Sortino} & \textbf{Max DD (bps)} & \textbf{Calmar} & \textbf{Win Rate} \\
\midrule
MSE (Baseline) & 0.27 & 0.35 & -52.16 & 0.29 & 0.54 \\ 
Huber & 0.32 & 0.43 & -52.84 & 0.34 & 0.55 \\ 
Gaussian NLL & 0.76 & 1.18 & -39.12 & 1.1 & 0.52 \\ 
Student-t NLL & -0.04 & -0.06 & -4932.33 & -0.05 & 0.52 \\ 
Quantile Loss & -0.46 & -0.66 & -72.09 & -0.36 & 0.48 \\ 
Adaptive Conformal (MSE) & -0.15 & -0.2 & -59.96 & -0.14 & 0.54 \\ 
\textbf{Evidential Regression (ProbFM core)} & 0.17 & 0.22 & -42.44 & 0.23 & 0.55 \\ 
\bottomrule
\end{tabular}}
\end{center}
\end{table}

\vspace{-4ex}
\begin{table}[H]
\caption{Trading Performance Comparison for SOL}
\vspace{-2ex}
\label{tab:trading_performance_sol}
\begin{center}
\resizebox{\linewidth}{!}{%
\begin{tabular}{lccccccc}
\toprule
\textbf{Method} & \textbf{Annual Sharpe} & \textbf{Annual Sortino} & \textbf{Max DD (bps)} & \textbf{Calmar} & \textbf{Win Rate} \\
\midrule
MSE (Baseline) & 1.02 & 1.51 & -45.41 & 1.62 & 0.54 \\ 
Huber & 1.07 & 1.59 & -45.41 & 1.69 & 0.52 \\ 
Gaussian NLL & 1.09 & 1.6 & -45.41 & 1.73 & 0.54 \\ 
Student-t NLL & 1.4 & 2.38 & -2899 & 3.48 & 0.51 \\ 
Quantile Loss & 0.81 & 1.21 & -45.41 & 1.29 & 0.51 \\ 
Adaptive Conformal (MSE) & 0.99 & 1.47 & -45.41 & 1.57 & 0.54 \\ 
\textbf{Evidential Regression (ProbFM core)} & 1.64 & 2.84 & -28.99 & 4.07 & 0.54 \\ 
\bottomrule
\end{tabular}}
\end{center}
\end{table}

\vspace{-4ex}
\begin{table}[H]
\caption{Trading Performance Comparison for USDC}
\vspace{-2ex}
\label{tab:trading_performance_usdc}
\begin{center}
\resizebox{\linewidth}{!}{%
\begin{tabular}{lccccccc}
\toprule
\textbf{Method} & \textbf{Annual Sharpe} & \textbf{Annual Sortino} & \textbf{Max DD (bps)} & \textbf{Calmar} & \textbf{Win Rate} \\
\midrule
MSE (Baseline) & 7.24 & 14.01 & -0.06 & 37.02 & 0.61 \\ 
Huber & 5.57 & 10.41 & -0.07 & 25.33 & 0.58 \\ 
Gaussian NLL & 3.5 & 7.02 & -0.07 & 16.47 & 0.51 \\ 
Student-t NLL & 8.4 & 14.85 & -6 & 41.74 & 0.65 \\ 
Quantile Loss & 0.19 & 0.32 & -0.12 & 0.52 & 0.43 \\ 
Adaptive Conformal (MSE) & 8.65 & 14.54 & -0.06 & 42.66 & 0.66 \\ 
\textbf{Evidential Regression (ProbFM core)} & 9.14 & 16.24 & -0.05 & 53.43 & 0.68 \\ \bottomrule
\end{tabular}}
\end{center}
\end{table}

\vspace{-4ex}
\begin{table}[H]
\caption{Trading Performance Comparison for USDT}
\vspace{-2ex}
\label{tab:trading_performance_usdt}
\begin{center}
\resizebox{\linewidth}{!}{%
\begin{tabular}{lccccccc}
\toprule
\textbf{Method} & \textbf{Annual Sharpe} & \textbf{Annual Sortino} & \textbf{Max DD (bps)} & \textbf{Calmar} & \textbf{Win Rate} \\
\midrule
MSE (Baseline) & 3.8 & 6.6 & -0.14 & 11.62 & 0.51 \\ 
Huber & 3.2 & 5.38 & -0.11 & 12.56 & 0.49 \\ 
Gaussian NLL & 0.91 & 1.46 & -0.21 & 1.91 & 0.44 \\ 
Student-t NLL & 4.78 & 8.03 & -11.01 & 18.31 & 0.55 \\ 
Quantile Loss & 0.09 & 0.16 & -0.25 & 0.16 & 0.4 \\ 
Adaptive Conformal (MSE) & 3.86 & 6.4 & -0.12 & 13.76 & 0.53 \\ 
\textbf{Evidential Regression (ProbFM core)} & 5.28 & 8.32 & -0.11 & 20.03 & 0.58 \\ 
\bottomrule
\end{tabular}}
\end{center}
\end{table}

\vspace{-4ex}
\begin{table}[H]
\caption{Trading Performance Comparison for XRP}
\vspace{-2ex}
\label{tab:trading_performance_xrp}
\begin{center}
\resizebox{\linewidth}{!}{%
\begin{tabular}{lccccccc}
\toprule
\textbf{Method} & \textbf{Annual Sharpe} & \textbf{Annual Sortino} & \textbf{Max DD (bps)} & \textbf{Calmar} & \textbf{Win Rate} \\
\midrule
MSE (Baseline) & 1.5 & 2.82 & -28.18 & 3.71 & 0.53 \\ 
Huber & 1.25 & 2.27 & -26.35 & 3.31 & 0.54 \\ 
Gaussian NLL & 1.07 & 2 & -28.68 & 2.6 & 0.52 \\ 
Student-t NLL & 1.12 & 2.12 & -2634.95 & 2.98 & 0.52 \\ 
Quantile Loss & 1.58 & 2.9 & -33.84 & 3.25 & 0.55 \\ 
Adaptive Conformal (MSE) & 1.22 & 2.27 & -28.35 & 3.01 & 0.52 \\ 
\textbf{Evidential Regression (ProbFM core)} & 1.44 & 2.67 & -28.35 & 3.55 & 0.53 \\ 
\bottomrule
\end{tabular}}
\end{center}
\end{table}

\section{Related Work}\label{appendix:related_work}
\input{5.relatedwork}

\section{Limitations}\label{appendix:future_work}

While our evaluation is comprehensive, several limitations warrant discussion as follows:

\begin{itemize}
\item \textbf{Dataset Scope}: Our primary evaluation focuses on eleven cryptocurrencies. Thus, broader evaluation across more assets and longer time periods would strengthen conclusions.
\item \textbf{Market Regime Coverage}: We understand that the evaluation period may not capture all possible market regimes (e.g., extreme crashes, prolonged bear markets), in which we will further cover more market regimes in subsequent works.
\item \textbf{Transaction Costs}: Our trading simulations do not account for slippage, fees, or market impact. Real-world implementation would require incorporating these factors.
\item \textbf{Alternative Architectures}: We focus on LSTM-based implementations. Other architectures 
may yield different trade-offs.
\end{itemize}

\section{Future Work}\label{appendix:future_work}
We outline several directions for future research that address current limitations and extend the applicability:
\begin{itemize}
    \item \textbf{Cross-Domain Generalization:} We plan to extend evaluation beyond cryptocurrency markets to validate generalizability across diversified domains. This includes financial markets (equities, foreign exchange, commodities) as well as non-financial time series applications such as energy demand forecasting, traffic prediction, and industrial sensor data. Such comprehensive evaluation will establish whether the epistemic-aleatoric decomposition benefits transfer across domains with fundamentally different noise characteristics and temporal dynamics.
    
    \item \textbf{Component-Level Ablation Studies:} To better understand the contribution of individual architectural components, we will conduct systematic ablation studies that isolate the effects of: (i) coverage loss integration, (ii) evidence annealing schedule, and (iii) bounded prediction transform. These ablations will provide clearer guidance on which design choices are essential for different application contexts.
    
    \item \textbf{Statistical Robustness Analysis:} 
    Future work will incorporate multiple random initializations per configuration
    with comprehensive statistical analysis including confidence intervals, standard deviations across runs, and formal hypothesis testing to quantify significance of performance differences. This is particularly important given the inherent noise in financial time series data.
    
    \item \textbf{Classical Baseline Comparisons:} To establish absolute performance benchmarks, we will include comparisons against classical time series and financial econometric models, including ARIMA, GARCH, and their variants. These baselines will provide context for understanding the value-add of deep learning approaches and help identify regimes where simpler models may suffice.
    
    \item \textbf{Multi-Horizon and Multivariate Extensions:} We plan to extend the framework to multi-horizon and multivariate forecasting using Normal-Inverse-Wishart priors, enabling joint uncertainty quantification across forecast horizons and correlated assets. 
\end{itemize}

%% file: 5.relatedwork.tex
The landscape of probabilistic time series forecasting has evolved significantly with the emergence of foundation models. Traditional approaches like DeepAR \citep{salinas2020deepar} established parametric probabilistic forecasting through autoregressive RNN architectures, while Temporal Fusion Transformers \citep{lim2021temporal} advanced the field with attention mechanisms and quantile-based uncertainty estimation.

Recent foundation models represent major advances: Lag-Llama \citep{rasul2023lagllama} demonstrates the effectiveness of Student's t-distribution loss in foundation model training, achieving superior performance through end-to-end probabilistic learning. TimeGPT \citep{garza2023timegpt} introduces conformal prediction to foundation models, providing coverage guarantees through non-parametric frameworks. Toto \citep{cohen2024toto} advances the field with sophisticated Student-T mixture models optimized for observability data, showing that mixture approaches can handle complex distributional characteristics. MOIRAI \citep{woo2024moirai} establishes universal forecasting capabilities through flexible four-component mixtures, demonstrating strong cross-domain performance.

DER emerges from the need for principled uncertainty quantification in neural networks \citep{sensoy2018evidential, amini2020deep}. Originally developed for classification tasks \citep{sensoy2018evidential} and later extended to 
regression \citep{amini2020deep}, DER learns higher-order distributions over distribution parameters, enabling explicit epistemic-aleatoric uncertainty decomposition. Our work represents the first comprehensive application of DER principles to time series forecasting with foundation model architectures.